\documentclass[runningheads]{llncs}
\usepackage[T1]{fontenc}
\usepackage{graphicx}
\usepackage{booktabs}
\usepackage[misc]{ifsym}

\usepackage{tikz}
\usetikzlibrary{arrows,shapes}

\tikzstyle{startstop} = [circle, draw, text centered, text width=50pt, fill=black!20!white]
\tikzstyle{process} = [rectangle, draw, minimum width=3cm, minimum height=1cm, text centered, text width=150pt]
\tikzstyle{decision} = [diamond, draw, aspect=2, text centered, text width=70pt]
\tikzstyle{arrow} = [thick,->,>=stealth]

\begin{document}

\title{Synthetic Non-stationary Data Streams for~Recognition of the Unknown}

\author{Joanna Komorniczak}

\institute{Wrocław University of Science and Technology, Wrocław, Poland
\email{joanna.komorniczak@pwr.edu.pl}}

\authorrunning{J. Komorniczak}

\maketitle              

\begin{abstract}
The problem of data non-stationarity is commonly addressed in~data stream processing. In~a~dynamic environment, methods should continuously be ready to~analyze time-varying data -- hence, they should enable incremental training and respond to~\emph{concept drifts}. An equally important variability typical for non-stationary data stream environments is the emergence of new, previously unknown classes. Often, methods focus on one of these two phenomena -- detection of \emph{concept drifts} or detection of \emph{novel classes} -- while both difficulties can be observed in~data streams. Additionally, concerning previously unknown observations, the topic of \emph{open set} of classes has become particularly important in~recent years, where the goal of methods is to~efficiently classify within known classes and recognize objects outside the model competence. This article presents a~strategy for synthetic data stream generation in~which both concept drifts and the emergence of new classes representing \emph{unknown} objects occur. The presented research shows how unsupervised drift detectors address the task of detecting novelty and concept drifts and demonstrates how the generated data streams can be utilized in~the \textit{open set recognition} task.

\end{abstract}

\section{Introduction}

Machine learning methods are used in~many real-world scenarios, such as medical diagnostics \cite{mendelson2020online}, device monitoring \cite{sahal2020big}, fraud or intrusion detection \cite{tavallaee2009detailed}, and analysis of content appearing on the web \cite{li2022event}, to~name a~few. Almost all of~these real-world applications involve data variability over processing time. This non-stationarity of data can manifest itself in~the form of the emergence of new diseases and drugs in~the case of diagnostic tasks, new fraud methods, or new trends in~social media.

\emph{Concept drifts} have been defined as changes in~the probabilistic distribution of the data visible over time \cite{gama2004learning}. Reacting to~such changes remains necessary due to~the decrease in~recognition quality when \emph{real} drifts occur. Any changes in~the distribution of the incoming data (including those described as~\emph{virtual} drifts -- which do not lead to~a~shift in~the decision boundary) may indicate important phenomena occurring in~the analyzed domain \cite{hoens2012learning}. One such phenomenon may be the appearance of unknown observations that form a~new category of objects.

\begin{figure}
    \centering
    \includegraphics[width=0.75\linewidth]{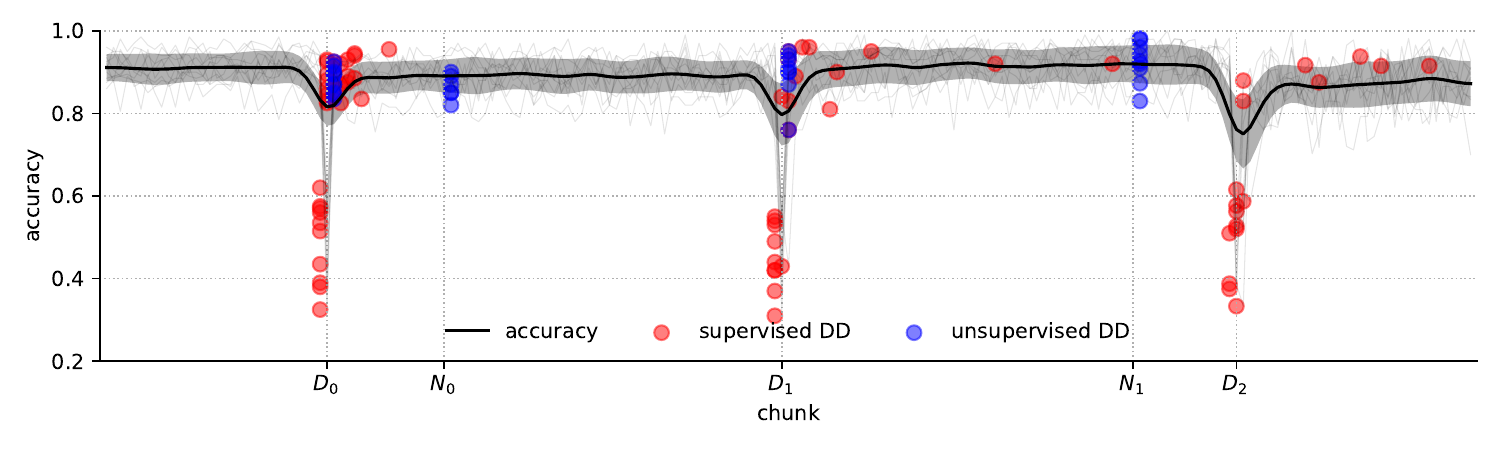}
    \caption{Drift detection moments signaled by supervised (red) and unsupervised (blue) drift detectors in~the non-stationary data stream with concept drifts in~\textsc{kc} (marked as~$D$) and the emergence of \textsc{uc} (marked as $N$).}
    \label{fig:adwin}
\end{figure}

Since the appearance of a~new class affects the data distribution, concept drift detectors should also allow for the recognition of such a~change. However, unknown objects will most often not receive labels, meaning that \emph{explicit} drift detectors -- those that monitor classification quality during processing -- will not allow the change detection \cite{gozuaccik2021concept}. Figure~\ref{fig:adwin} shows the effect of detecting concept drifts using the supervised \textit{Adaptive Windowing} detector \cite{bifet2007learning} and unsupervised \textit{Centroid Distance Drift Detector} \cite{klikowski2022concept}. The horizontal axis shows the moments of concept drift in~known classes (\textsc{kc}), for which labels are provided, and the emergence of new, unknown classes (\textsc{uc}). Only unsupervised methods allow for detecting the concept change related to~novelty, assuming no inflow of labels for unknown observations before the new class is detected and, therefore, no possibility of~estimating the recognition quality of the classifier.

The drift detection methods are most often evaluated with an assumption of~no new class appearance. Meanwhile, \textit{novelty detection} forms a~different task, where the methods aim to~recognize the appearance of a~group of samples describing a~set of new observations, separate from the previously analyzed classes \cite{faria2016novelty}. Furthermore, canonical classification methods that recognize within a~\emph{closed set} of classes will assign objects from \emph{unknown} classes to~\emph{known} categories, which were used to~train the classification model \cite{scheirer2012toward}. Such behavior of systems can lead to~significant errors in~their operation or allow misleading the recognition model. It should be noted here that (a) the occurrence of concept drifts within \textsc{kc}, (b) the emergence of new classes, and (c) the design of robust classifiers are consistently associated with incremental non-stationarity. Meanwhile, in~the literature, these issues are often treated separately, and methods dedicated exclusively to~one of the described difficulties are proposed.

With the noticeable lack of real-world data streams in~the scientific community that would be characterized by various types of non-stationarity -- such as~concept drifts and observations of new, unknown classes -- the proposed algorithms are often evaluated on synthetic streams \cite{bifet2017classifier}. The \emph{concept drift ground truth} -- which describes the exact moments where the changes in~the data distribution occur -- makes the use of synthetic data necessary for reliable drift detection evaluation. In~the case of synthetic data, a~significant advantage is the possibility of a~precise specification of stream parameters, such as dimensionality, length, number of drifts, and characteristics of classes. This allows studying the performance of methods from diverse angles and enables statistical analysis of the results obtained in~individual experiment replications.

\paragraph{Contribution}

This work proposes and describes an \textit{Open World Data Stream Generator with Concept Non-stationarity} (\textsc{owdsg}), in~which static classes are generated using the \textit{Madelon} generator \cite{guyon2003design}, and then arranged into a~continuous data stream that can be characterized by both concept drifts and the emergence of new classes. The implementation of the method in~\textit{python} is publicly available on the GitHub repository \footnote{\url{https://github.com/JKomorniczak/OWDSG}}.

In addition to~the description of the generator itself and the presentation of the generated streams, this work presents two experiments: (a) unsupervised detection of \emph{concept drifts} and \emph{novelty} for different sensitivities of drift detectors, and (b) \emph{open set recognition} (\textsc{osr}) with incremental learning of new classes during processing.

The first experiment verifies whether the novelty detection task can be effectively performed by concept drift detectors with appropriate threshold selection. The second experiment examines the quality of classification and recognition of unknown samples in~the experimental protocol that provides labels for the initially unknown classes. In~the \textsc{osr} task, recognition quality was examined for the \textit{Multilayer Perceptron} (\textsc{mlp}) classifier with decision support thresholding.

\section{Related works}

The presented generation method can be used in~several areas related to~non-stationary data processing: \emph{concept drift detection}, \emph{novelty detection}, and \emph{open-set recognition} in~a~streaming environment.

\subsection{Concept drift detection}

The available studies have shown that the evaluation of drift detection using classification quality causes a~bias towards overactive methods \cite{bifet2017classifier}. For detection quality evaluation, synthetic data streams are often used, where the moments of~concept drift occurrence are predetermined and known. Particularly popular are the generators available in~\textit{MOA}, \textit{River}, and \textit{scikit-multiflow} packages, namely SEA, LED, and Hyperplane \cite{gulcan2023unsupervised,abbasi2021elstream,korycki2021concept,ali2025novel,lukats2025benchmark}. Among other popular generators, one can mention the generator available in~the \textit{stream-learn} library that uses the \textit{Madelon} generator as a~baseline. The streams generated with the library are used in~the evaluation of drift detection \cite{chiang2024detection} and classification of imbalanced and non-stationary data streams \cite{zyblewski2021preprocessed}. Some research uses streams in~which drift manifests itself in~varying statistical properties of data \cite{liu2018accumulating} or employs generators that focus on class imbalance \cite{brzezinski2021impact}.

\subsection{Novelty detection}

In \textit{novelty detection} research, both synthetic and real-world datasets are used. The utility of real-world data comes from the possibility of assessing methods based on classification quality metrics \cite{faria2016novelty}. Among popular real-world data, researchers often evaluate their models on \textit{KDD Cup 99} \cite{tavallaee2009detailed}, \textit{COVID-19} \cite{mendelson2020online}, \textit{ForestCoverType} \cite{de2016minas,pevny2016loda} and \textit{Shuttle} \cite{tan2011fast}. For the synthetic data stream, similarly to~the task of concept drift detection, often generators from \textit{MOA} and \textit{scikit-multiflow} are employed, particularly focusing on \textit{RandomRBF} generator \cite{gaudreault2025prioritizing,din2021data}. 

Although the lack of consideration of concept variability in~novelty detection research is a~frequently addressed research gap \cite{faria2016novelty}, some works consider both phenomena. In~the experiments evaluating \emph{Cluster Drift Detection} \cite{mendelson2020online} the authors used two-dimensional synthetic streams generated based on the \verb|make_blobs| function available in~the \textit{scikit-learn} library. Another method \cite{masud2011detecting} focuses on class recurrence as a~type of~concept evolution in~the context of novelty discovery. The experiments presented in~the mentioned research used synthetic data streams generated using \textit{SynC} generator \cite{masud2010classification}. This generation method creates data with concept drifts only, where features are specified based on rotating hyperplanes. The publication also presented \textit{SynNC} generator that allows for simulating the changes in~the distribution of known classes (concept drifts) as well as the emergence of novel classes. The objects in~\textit{SynNC} are sampled from Gaussian distributions of different means and variances, and the drifts are simulated by shifting the means of the probability distributions.

\subsection{Open set recognition}

\textit{Open set recognition} is most often considering static data, where real-world multiclass datasets are used \cite{scheirer2012toward}. Since \textsc{osr} frequently addresses the limitations and vulnerabilities of non-explainable models such as neural networks \cite{geng2020recent}, the research uses mostly non-tabular image or text data. Despite the popular assumption of methods being employed in~the static and stationary environment, some research has brought the focus of \textsc{osr} into the area of data stream processing, emphasizing \emph{open environment challenges} that include the presence of concept drifts and anomalies \cite{diao2023oebench}.

One of the first methods designed for \textsc{osr} in~data streams, namely \textit{OSDe-SVM} \cite{lopez2022incremental} was considering the incremental learning environment of face recognition. This research used real-world image data. Another recent publication proposed a~framework for \textsc{osr} in~data streams \cite{barcina2024resilience} that combines incremental classification with incremental clustering. The research was carried out on synthetic datasets with a~fixed number of known and unknown classes and with the highest dimensionality of 12 features. The two types of data streams used in~the experiments described Gaussian clouds or were generated using the \textit{Madelon}. 

The commonness of synthetic data generated with \textit{Madelon} shows the utility of this type of data in~non-stationary streaming scenarios. The use of the proposed \textsc{owdsg} method could allow for an extension of the research with a~more diverse cardinality of unknown classes, the problem dimensionality, or even a~continuous increase of the \textsc{uc} percentage.

\section{Open World Data Stream Generator with Concept Non-stationarity}

This section presents the \textit{Open World Data Stream Generator with Concept Non-stationarity} (\textsc{owdsg}). The method uses class distributions sampled with the \textit{Madelon} generator \cite{guyon2003design}, available in~the \emph{scikit-learn} library. The static samples are organized into a~data stream, where, depending on the hyperparameters, concept drifts in~\textsc{kc} occur and novel, \textsc{uc} emerge. The generation method will optionally apply feature projection if the dimensionality requirements of the baseline generator are not met.

\begin{table}[!b]
    \centering
    \scriptsize
    \caption{Description of the \textsc{owdsg} hyperparameters and their default values. Hyperparameters with an asterisk (*) are propagated to~the underlying static data generator.}
    \setlength{\tabcolsep}{8pt}

    \begin{tabular}{p{3.5cm}|p{7.5cm}}
    \toprule
         \textsc{hyperparameter} & \textsc{description} \\
    \midrule
        \verb|n_chunks = 200| & Length of the data stream in~batches.  \\
        \verb|chunk_size = 200| & Number of problem samples in~each batch. \\
    \midrule
        \verb|n_drifts| & Number of concept drifts in~the stream $n_d$. \\
        \verb|n_novel| & Number of new classes in~the stream $n_n$. \\
        \verb|percentage_novel = 0.1| & The proportion of a~chunk that will be populated with novel class samples. \\
        \verb|even_gt = True| & The boolean value describing if the concept drifts and the moments on new class novelties are equally distributed across the stream. \\
        \verb|hide_label = False| & The boolean value describing if labels of all unknown classes are unified to~a~common value.\\
    \midrule
        \verb|n_classes = 2| & Number of known classes $n_c$. \\
        \verb|weights = [0.5, 0.5]| & Proportions of known classes. \\
        \verb|n_clusters_per_class = 1|* & Number of clusters in~each class.\\
        \verb|class_sep = 1|* & The side lengths of a~hypercube for data generation. \\
        \verb|n_features = 10|* & Dimensionality of the data. \\
        \verb|n_informative = 10|* & Number of informative features.\\
    \midrule
        \verb|allow_projection = True| & The boolean value determining if the random feature projections can be performed. If the value is False, the generation process will result in~an error when prerequisites are not satisfied. Otherwise, the method will perform projections to~a~requested dimensionality. When projections are performed, the increase of \verb|class_sep| value is recommended.\\
        \verb|random_state = None|* & Random seed for data generation. \\
    \bottomrule
    \end{tabular}
    \label{tab:gen-hyperparams}
\end{table}

In the original \textit{Madelon}, the number of generated class clusters $N$ must be smaller than $2^d$, where $d$ describes the number of informative attributes of~the problem. When generating data streams with concept drifts and new classes, the generator will generate $N = (n_d \cdot n_c) + n_n$ classes assuming a~single cluster per class, where $n_{d}$ describes the number of~drifts in~\textsc{kc}, $n_c$ the number of \textsc{kc} and $n_n$ the number of novel, unknown classes. In~many cases, the use of~the original generator forces high dimensionality of the problem. The proposed \textsc{owdsg} allows for the generation of data streams with the requested number of features employing random projections, reducing the data to~a~lower-dimensional feature space. The generator hyperparameters are described in~Table~\ref{tab:gen-hyperparams}. 

\begin{figure}[!b]
		\centering
    	\resizebox{0.7\textwidth}{!}{\begin{tikzpicture}[font=\sffamily, text centered]

\node (start) at (0,0) [startstop] {Get Parameters};
\node (define) at (0,-2) [process] {Define Concept Drift and Novelty Ground Truth};
\node (decision) at (0,-4) [decision] {Dimensionality correct};

\node (n1) at (-4,-5.5) [process] {Correct dimensionality and number of informative features};
\node (n2) at (-4,-7) [process, fill=blue!20!white] {Generate data with Madelon with modified hyperparameters};
\node (n3) at (-4,-8.5) [process, fill=red!20!white] {Perform random feature projections to requested dimensionality};

\node (y1) at (4,-7) [process, fill=blue!20!white] {Generate data with Madelon};

\node (decision2) at (0,-11) [decision] {Length of the stream reached};

\node (n4) at (-4,-12.5) [process] {Establish label of KC and UC for sampling};
\node (n5) at (-4,-14) [process, fill=red!20!white] {Sample unused KC into current chunk with \textit{weight} parameter};
\node (n6) at (-4,-15.5) [process, fill=red!20!white] {Replace KC samples with UC with \textit{percentage novel} parameter};
\node (n7) at (-4,-17) [process] {Shuffle chunk and add to data stream};

\node (decision3) at (4,-12.5) [decision] {Hide index};

\node (y2) at (4,-14.75) [process] {Change label of all UC to common value};
\node (stop) at (4,-17) [startstop] {Return data stream};

\draw [arrow] (start) -- (define);
\draw [arrow] (define) -- (decision);
\draw [arrow] (decision.west) -- node[above left] {N} (-4,-4) -- (n1.north);
\draw [arrow] (decision.east) -- node[above right] {Y} (4,-4) -- (y1.north);
\draw [arrow] (n1) -- (n2);
\draw [arrow] (n2) -- (n3);

\draw [arrow] (n3.south) -- (-4,-9.5) -- (0,-9.5) -- (decision2.north);
\draw [arrow] (y1.south) -- (4,-9.5) --  (0,-9.5) -- (decision2.north);

\draw [arrow] (decision2.west) -- node[above left] {N} (-4,-11) -- (n4.north);
\draw [arrow] (n4) -- (n5);
\draw [arrow] (n5) -- (n6);
\draw [arrow] (n6) -- (n7);
\draw [arrow] (n7) -- (0, -17) -- (decision2.south);

\draw [arrow] (decision2.east) -- node[above left] {Y} (4,-11) -- (decision3.north);
\draw [arrow] (decision3) -- node[right] {Y} (y2);
\draw [arrow] (y2) -- (stop);

\draw [arrow] (decision3.west) -- node[above right] {N} (0.5,-12.5) -- (0.5,-17) -- (stop.west);

\end{tikzpicture}}
		\caption{The scheme of \textit{Open World Data Stream Generator with Concept Non-stationarity} method, employing the \textit{Madelon} static generation method (blue blocks) and organizing samples into a~data stream with requested dimensionality (red blocks).}
		\label{fig:generator-scheme}
	\end{figure}
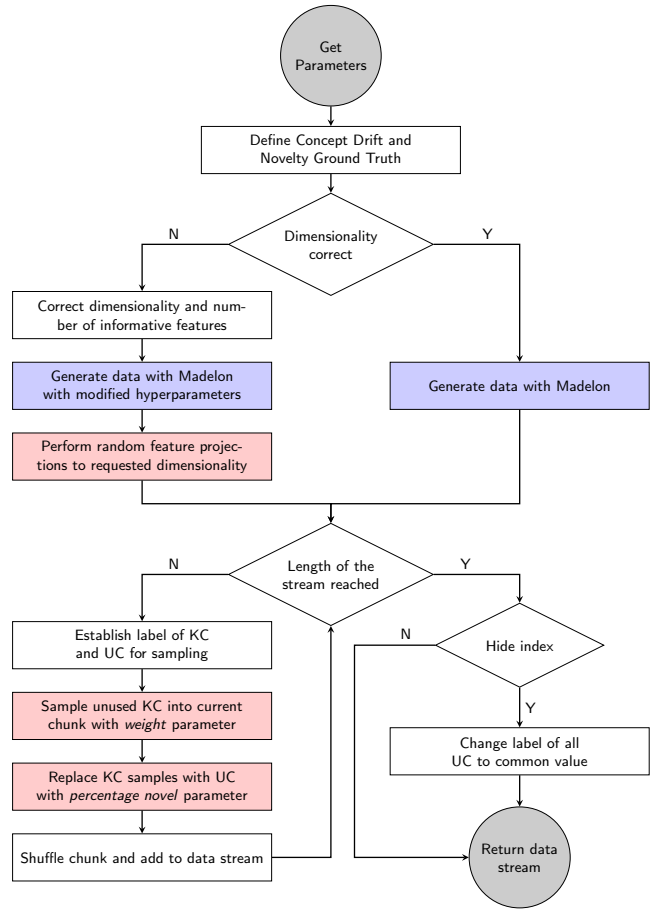

The scheme of the \textsc{owdsg} method is presented in~Figure~\ref{fig:generator-scheme}. At~the beginning of the processing, depending on the stream length and the number of events (drifts and class novelties), the method determines the chunks in~which they will appear. The indices of chunks in~which events will occur are described as \emph{ground truth} and are determined independently. Then the method checks whether the passed hyperparameters describing the dimensionality of the problem allow the use of the original \textit{Madelon} generator. Only in~the case when the desired dimensionality is too low and the hyperparameter \verb|allow_projection| permits it, the dimensionality of the generated data is changed, and then the method performs random projections to~the desired number of features.

In the following steps, the generated data is arranged into a~stream. If the stream length has not yet been reached, another chunk is created. The samples belonging to~the processed data batch are drawn from the pool of static samples. Concept drift causes a~change of the clusters from which \textsc{kc} are sampled, but does not affect the distribution of \textsc{uc}. As a~result, only classes known at~the beginning of processing will manifest concept drifts. This makes the drift detection task more challenging in~the further state of data processing -- as some distributions (\textsc{uc}) remain stable despite the occurrence of \textsc{kc} drifts. In~the object selection procedure for a~specific data batch, \textsc{kc} samples are drawn first. Cardinality of individual classes will depend on the \verb|weights| parameter indicating the proportion between the known classes. In~the next step, samples for each \textsc{uc} replace the samples from \textsc{kc}. In~the presented generation procedure, both concept drifts and the appearance of unknown classes are abrupt. As the final step of the generation procedure, the labels of the \textsc{uc} samples can be optionally changed to~a~common one, equal to~$n_c$. 

\begin{figure}[!htb]
    \centering
    \includegraphics[width=0.95\linewidth]{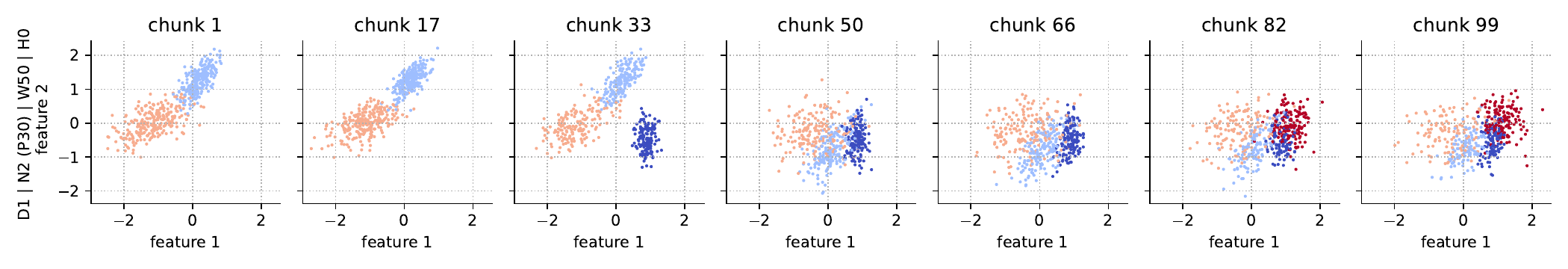}
    \includegraphics[width=0.95\linewidth]{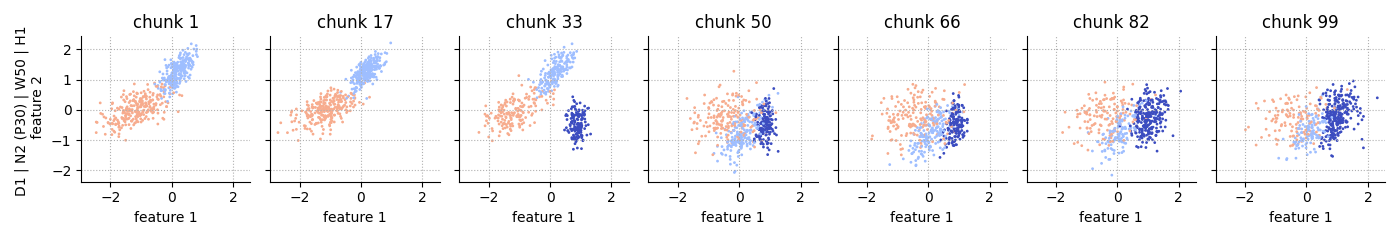}
    \caption{Exemplary planar data streams with a~single concept drift and two novel classes. The top row shows the original labels of \textsc{uc}, while the bottom row uses a~common label for all \textsc{uc}.}
    \label{fig:example-hide}
\end{figure}

Examples of two-dimensional data generated by \textsc{owdsg} are shown in~Figure~\ref{fig:example-hide} with visible and hidden labels of new classes. The projection mechanism available in~the generation procedure was used to~generate multi-class data (taking into account \textsc{uc}) of low dimensionality. The subsequent subplots visible in~the figures present seven data batches from a~stream of 100 chunks. The color of the point indicates its class.

The presented streams have a~single concept drift and two new class emergence moments. It can be seen that drift affects only \textsc{kc}, which can indicate difficulty for drift detection methods in~the later phase of stream processing. Streams with a~unified label, shown at~the bottom of the figure, can be used for \textsc{osr} experiments when classifiers are not trained using individual \textsc{uc} or can be used to~simulate an environment where the share of \textsc{uc} increases with processing time.

\begin{figure}[!htb]
    \centering
    \includegraphics[trim={0.7cm 0 1cm 0}, clip, width=0.47\linewidth]{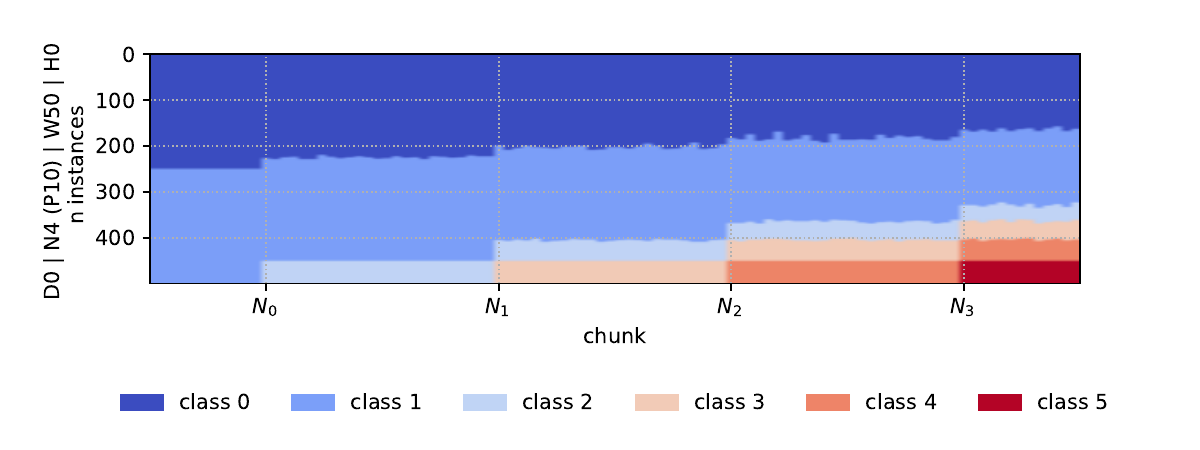}
    \includegraphics[trim={0.7cm 0 1cm 0}, clip, width=0.47\linewidth]{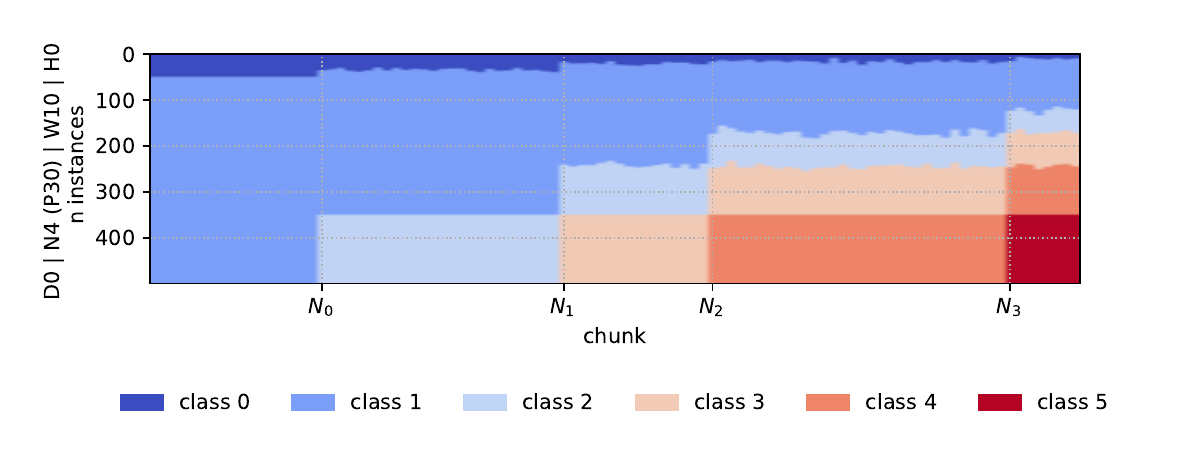}
    \caption{The class proportions in~the generated streams in~the context of novelty ground truth -- evenly distributed on the left and random on the right side -- and \textsc{kc} imbalance -- balanced on the left and imbalanced on the right side.}
    \label{fig:example-priors}
\end{figure}

Figure~\ref{fig:example-priors} shows how the proportions between classes change when a~new class appears. The horizontal axis of each plot shows subsequent chunks, and the colors indicate the number of instances of a~given class at~a~specific batch. Classes labeled 0 and 1 constitute the \textsc{kc} set. The horizontal axis indicates the moments when \textsc{uc} classes appear. There are no concept drifts in~the presented streams, as they do not affect the proportions between classes.

The plots show streams generated by different \textsc{owdsg} configurations. The~left part of the figure shows the data stream with four evenly distributed \textsc{uc} emergence moments and balanced \textsc{kc}. In~contrast, the right part shows a~stream where moments of appearance of new classes are determined randomly. In~the second stream, the proportions between \textsc{kc} have been modified -- the minority class constitutes 10\% of observations. The samples of the latest \textsc{uc} (the one that appeared last) constitute the ratio of observations specified by the \verb|percentage_novel|. As a~result of the random replacement of samples, this slightly reduces the number of samples from other classes.

The presented generation method aims to~enable and facilitate the experimental evaluation of the methods. The possibility of specifying many hyperparameters of the generator allows analyzing methods for \textit{concept drift detection}, \textit{novelty detection}, and \textit{open set recognition} under various conditions. \textsc{Owdsg} enables the simulation of concept drifts, which are relatively rarely taken into account in~novelty detection and open set recognition research but remain important from the perspective of real-world systems \cite{diao2023oebench}.

\section{Experimental design}

This publication, in~addition to~the data stream generation method, presents the results of two experiments for \emph{concept drift detection} and \emph{open set recognition} tasks, with the second assuming the incremental training of the classification model.

\subsection{Goals of experiments}

\paragraph{Concept drift detection}
The first experiment evaluated unsupervised drift detectors, focusing on their ability to~distinguish concept drifts from the emergence of~new classes, depending on the method's sensitivity to~changes. The evaluated detectors and their configuration are presented in~Table~\ref{tab:det-hyperparams}. For each of the detectors in~the sensitivity range, 10 values of the parameter were evenly sampled.

\begin{table}[]
    \centering
    \scriptsize
    \caption{The description of evaluated unsupervised drift detectors and their selected hyperparameters.}
    \setlength{\tabcolsep}{8pt}

    \begin{tabular}{p{1cm}|p{3.5cm}|p{6cm}}
    \toprule
    \textsc{acronym} & \textsc{name} & \textsc{evaluated hyperparameters} \\
    \midrule
    \textsc{md3} & Margin Density Drift Detector \cite{sethi2015don} & \verb|sigma| value sampled from the range $[0.1, 0.45]$\\
    \textsc{cddd} & Centroid Distance Drift Detector \cite{klikowski2022concept} & \verb|percent| hyperparameter sampled from the range $[0.6, 0.95]$\\
    \textsc{ocdd} & One Class Drift Detector \cite{gozuaccik2021concept} & the original implementation was modified to~enable \verb|sensitivity| thresholding in~the range $[0.3, 2.5]$\\
    \textsc{padd} & Parallel Activations Drift Detector \cite{komorniczak2024unsupervised} & \verb|threshold| parameter sampled from the range $[0.05, 0.5]$\\
    \bottomrule
    \end{tabular}
    \label{tab:det-hyperparams}
\end{table}

This experiment focuses on unsupervised detectors, assuming that in~a~dynamic, real-world environment, labels for unknown classes will not arrive without prior identification of unknown classes. The experiment tries to~verify whether setting an appropriate threshold (lower than in~the case of concept drift detection) will allow for the detection of both drifts within known classes and class novelty.

\paragraph{Open Set Recognition}

The next experiment examined the recognition quality when unknown categories emerge -- where some of the new classes were used to~incrementally train the model. In~this processing scenario, the last of \textsc{uc} always remains unknown. At~first, the number of known classes used to~train the classifier is equal to~$n_c$. When the first new class appears, its labels remain hidden, and the model is trained only using \textsc{kc}. Later, when another \textsc{uc} appears, the labels for all previously visible ones are provided. Such a~processing scheme will make objects from the newest class unknown to~the model. 

This experiment follows the standard \emph{test-then-train} protocol for \textsc{kc} and identified \textsc{uc}, meaning that the sample’s labels arrive in~the next processing step. However, it should be noted that in~real-world applications labeling is~a~costly procedure that requires expert assistance, and hence the inflow of labels is almost always delayed \cite{grzenda2020delayed}.

The \textsc{osr} experiment investigates the performance of an \textsc{mlp} classifier with 100 neurons in~a~single hidden layer, which was trained for 10 epochs with each data chunk. The classifier was initialized with the number of outputs equal to~$n_c + n_n$, which allowed incremental training without the need to~rebuild the model. At~a~given processing point, only as many outputs of the classifier were considered as the classes for which labels arrived.

For illustration purposes, the experiment investigated the performance of~the model with six different threshold values $\theta$ to~detect the unknown samples. The threshold selection was based on the mean $\mu$ and standard deviation $\sigma$ of the classifier's support for the training samples. The threshold is calculated as: $\theta = \mu - \epsilon \cdot \sigma$. A~sample was assigned to~\textsc{kc} if the maximum support for a~given decision was greater than a~given threshold $y_{max}(x) > \theta$. This would indicate the assignment to~\textsc{kc} for samples with high decision \emph{certainty}. Otherwise, the sample was marked as unknown. The experiment evaluated six different values of $\epsilon$ in~a~range $[-0.5, 3]$. At~high values of $\epsilon$, more samples will be marked as known. In~contrast, at~low values of the parameter, only decisions with very high support will be made in~favor of known classes. In~extreme cases, all objects will be marked as \textsc{kc} (high $\epsilon$) or all marked as \textsc{uc} (low $\epsilon$).

In the presented \textsc{osr} experiment, certain assumptions were implemented, in~particular regarding the immediate inflow of labels. Simulating real evaluation conditions is a~complex task, not directly related to~the data generator but to~the experimental protocol, where one can, for example, introduce an additional factor delaying the appearance of labels by a~specified number of chunks or transfer labels for unknown classes only after previously detecting the novelty. The stream generator proposed in~this publication can be used in~such protocols for an even more reliable imitation of real-world non-stationary streaming environments.

\subsection{Evaluation metrics}

In the case of \textit{concept drift detection}, one of the valuable tools to~assess the quality of the method is visual analysis of the detection moments. The available numerical metrics allow estimating the number of incorrect drift signals and their delay \cite{bifet2017classifier} or the distances between detections and actual drifts \cite{komorniczak2024unsupervised}. Since the goal of the experiments is illustrative, the drift detection moments will be presented visually, allowing for a~straightforward assessment of the detector's sensitivity and localization of redundant detections.

Due to~the difficulties of assessing \textsc{osr} in~the context of class imbalance, the used measures are based on the \emph{balanced accuracy} metric. The use of this base metric is recommended when an~imbalance in~the context of the \textsc{kc}/\textsc{uc} dichotomy is observed or when the \textsc{uc} cardinality significantly exceeds the individual \textsc{kc} cardinality. With a selected metric, the quality was assessed using four different measures \cite{komorniczak2025taking}: 
\begin{itemize}
    \item \textit{outer score} -- describing the quality of binary recognition between \textsc{kc} and \textsc{uc};
    \item \textit{halfpoint score} -- estimating the quality of classification among \textsc{kc}, where the classifier is penalized for the incorrect assignment of \textsc{kc} towards \textsc{uc};
    \item \textit{inner score} -- canonical closed-set classification quality, where samples initially labeled as \textsc{uc} are assigned to~the most probable class among \textsc{kc};
    \item \textit{overall score} -- treating objects of \textsc{uc} as~a~separate class equivalent to~\textsc{kc}.
\end{itemize}

\begin{figure}[!htb]
    \centering
    \includegraphics[width=0.9\linewidth]{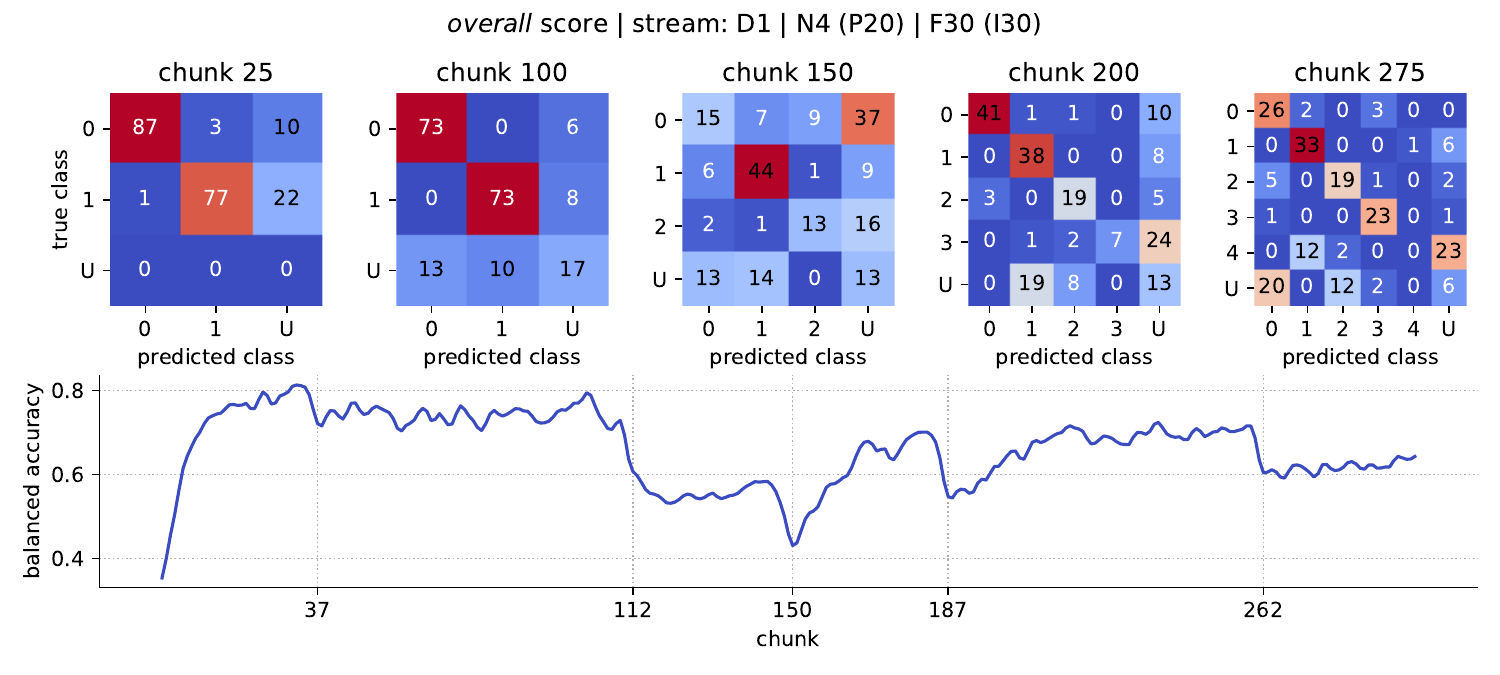}
    \caption{The \textit{overall score} confusion matrices in~specific chunks of the stream (top) and the plot presenting the metric value across the entire stream with a~single concept drift and four novel classes.}
    \label{fig:overall}
\end{figure}

Evaluation of the quality of \textit{open set recognition} is a~complex task. For~an~illustration, the \textit{overall score} computed throughout the stream processing was presented in~Figure~\ref{fig:overall}. The first row of the figure presents confusion matrices at~specific time instances. The last label describes the unknown class. In~the second row of the figure, the \textit{overall score} calculated in~each data chunk is presented. The evaluated data stream had a~single concept drift in~the 150th data chunk and four novel classes, whose appearance moments are also marked on the horizontal axis.

\subsection{Generated data streams}

Data streams with various characteristics were generated to~examine the behavior of the methods. For the first experiment, the focus was placed on the \verb|percentage_novel| hyperparameter indicating the share of \textsc{uc} in~subsequent chunks. The second experiment focused on the \verb|n_clusters_per_class| hyperparameter, as its increase should make the \textsc{osr} task more challenging.

For the first experiment, streams with evenly distributed 2 concept drifts and 3 new classes were generated. The data were five-dimensional so that drift detection would not pose a~significant challenge for the examined methods. New classes constituted 10\%, 20\%, and 50\% of observations in~chunks. 

For the second experiment, higher-dimensional data was generated -- samples were described by 70 informative features. Objects of new classes constituted 20\% of the batches. In~these streams, there was one concept drift at~the central point and four novel class appearances, evenly distributed over the processing time. The number of clusters in~each class was 1, 3, or 5.

For both experiments, the streams consisted of 300 chunks, each containing 500 objects. At~the beginning of the processing, only two classes were known. For each configuration, ten experiment replications were performed for different random seeds.

\section{Experimental evaluation}

This section presents and analyzes the results of the conducted experiments, showing the utility of the presented \textsc{owdsg}.

\subsection{Drift and novel class detection}

\begin{figure}[!htb]
    \centering
    \includegraphics[width=0.95\linewidth]{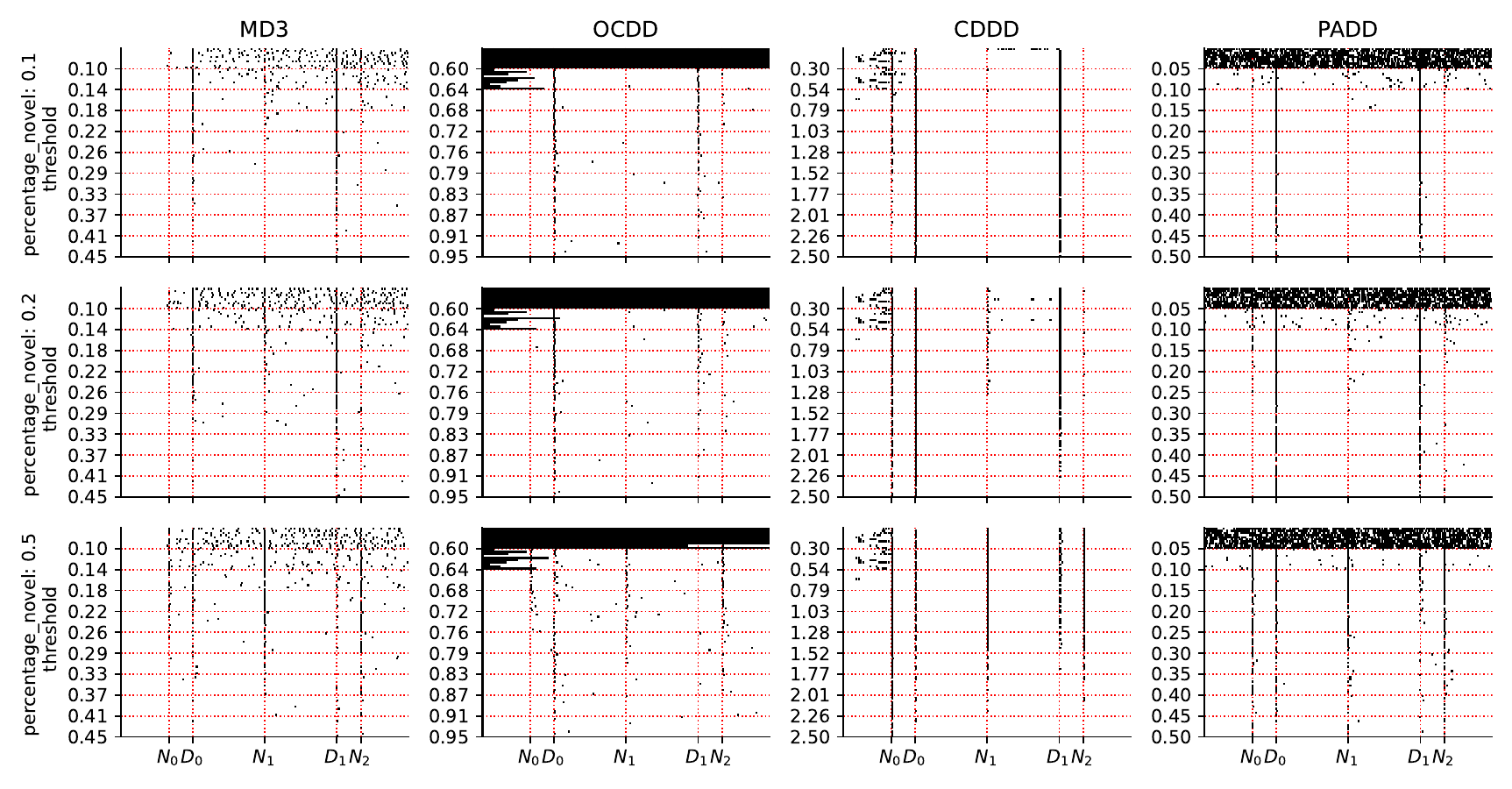}
    \caption{The results of the drift detection experiment for four drift detections (in columns) with various sensitivity in~the data streams with different proportions between \textsc{kc} and \textsc{uc} (in rows). The detection moments are marked with black points and the true events -- with red vertical lines.}
    \label{fig:dd-res}
\end{figure}

The first experiment examined the ability of four unsupervised concept drift detectors to~recognize drifts in~\textsc{kc} and the emergence of \textsc{uc}. The results for three different values of the \verb|percentage_novel| hyperparameter are shown in~Figure~\ref{fig:dd-res}, presenting the detectors in~columns and the share of \textsc{uc} in~rows. Each black segment in~the plot describes a~single detection within 10 replications for a~given stream and within 10 values of the hyperparameter indicating the detector's sensitivity. The sensitivity values are placed on the vertical axis of the subplots -- from the most sensitive at~the top to~the least sensitive at~the bottom. The horizontal axis and the vertical red lines indicate the moments of concept drifts (marked $D$) and the emergence of new classes (marked as $N$). A~detector that ideally recognizes concept drifts and novelties in~all replications would show the vertical black lines overlapping the red markers. Each single element lying far from the ticks is, therefore, a~redundant or late detection.

Generated streams were not intended to~pose a~challenge to~the evaluated drift detection methods. This experiment aimed to~inspect whether controlling the sensitivity hyperparameter would allow signaling only drifts in~the concept or both changes in~\textsc{kc} and the emergence of new classes. It can be seen that for a~small percentage of unknown classes (10\% in~the first row) even for a~very high sensitivity of the detectors, not all of them recognized the changes. Only the \textsc{cddd} detector marked the first \textsc{uc} emergence, but later it was not able to~detect the second and third \textsc{uc}. When increasing the proportion of \textsc{uc} to~50\%, all methods responded to~both concept drifts and novelty. There are visible configurations in~which the detectors will indicate concept drifts within the \textsc{kc}, but will not signal new classes. However, the selection of such a~universal pair of hyperparameters is a~challenge, and such calibration should be made precisely for a~specific type of stream and a~specific detection method.

\subsection{Open set recognition in~incremental learning setting}

\begin{figure}[!htb]
    \centering
    \includegraphics[width=0.95\linewidth]{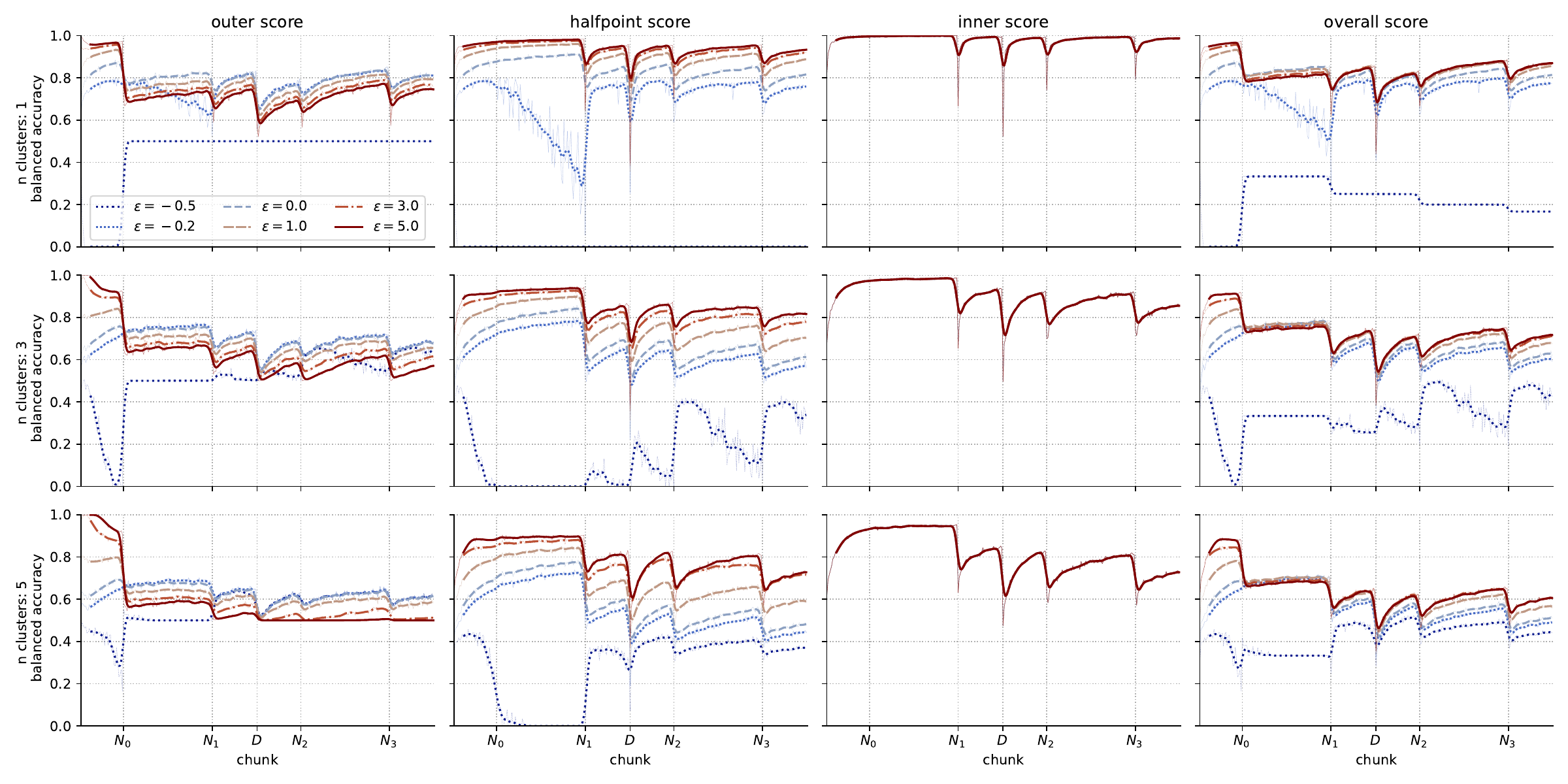}
    \caption{The results of \textsc{osr} experiment for four evaluation metrics (columns) and different numbers of clusters in~class (rows). The line colors indicate the $\epsilon$ hyperparameter value used to~calculate the recognition threshold.}
    \label{fig:exp-osr}
\end{figure}

The second experiment concerned the \textsc{osr}, where the most recent of the new classes was unknown, and the remaining ones were incrementally used to~train the model. The results of this experiment are presented in~Figure~\ref{fig:exp-osr} for different numbers of clusters per class in~rows and the four studied metrics in~columns. The colors of the individual plots indicate the values of the parameter $\epsilon$ used to~calculate the support threshold of the classifier.

The presented experiment aimed only to~demonstrate how the developed generator can be used to~evaluate \textsc{osr} methods and was not intended to~select the best recognition approach. However, based on the \textit{outer score} results, it can be seen that depending on the $\epsilon$, some methods achieved better results in~the identification of objects of unknown classes, reaching up to~80\% balanced accuracy. It can also be seen how the first appearance of the unknown class (marked $N_0$) significantly lowered the \textit{outer} and \textit{overall} scores. At~the time point marked as $N_1$, the second unknown class appeared, which means that labels for the previously visible new class started to~inflow. From the moment $N_1$, objects of this class were also used to~calculate the \emph{inner} and \textit{halfpoint score}, which is visible in~the drop in~quality at~this timestamp. A~single concept drift marked with a~$D$ marker negatively affected all metrics.

The second experiment showed how the recognition quality changes in~the context of class novelty and concept drift by evaluating the results of incremental training using previously unknown classes of the baseline \textsc{mlp} classifier with support thresholding.

\section{Conclusions}

This work describes a~strategy of non-stationary synthetic data stream generation in~which \emph{concept drifts} and \emph{novel classes} appear. The proposed \textsc{owdsg} approach uses the \textit{Madelon} generator, modifying its parameters and arranging samples into data streams of the desired dimensionality, length, and other characteristics specified by the end user. The data streams can represent concept drifts within \emph{known classes} and describe emerging novel, \emph{unknown classes}. Moreover, the generator allows for specifying the class imbalance and determining the proportion of unknown instances in~relation to~the number of known samples. To~generate data of low dimensionality, the generator uses random feature projections, which enable visualizing generated data streams in~planar space and conducting illustrative experiments.

The article additionally presents a~visualization of the obtained streams and performs two experiments focusing on (a) \emph{concept drift detection} task using four unsupervised approaches, and (b) \textit{open set recognition} task with incremental training using new classes.

The proposed generator can become a~valuable tool enabling the evaluation of methods dedicated to~processing data streams in~changing non-stationary environments, including taking into account the phenomena of concept drift and the emergence of new classes.

\begin{credits}
\subsubsection{\ackname} This work was supported by the~statutory funds of~the~Department of~Systems and~Computer Networks, Faculty of~Information and~Communication Technology, Wrocław University of~Science and~Technology

\subsubsection{\discintname}
The authors have no competing interests to~declare that are
relevant to~the content of this article.
\end{credits}

\bibliographystyle{splncs04}
\bibliography{bib.bib}

\end{document}